\documentclass[twoside]{article}
\usepackage{doublespace}
\usepackage{graphics,graphicx,amsmath}
\usepackage{cl}
\title{A Statistical Model for Word Discovery in Transcribed Speech}
\author{Anand Venkataraman\\STAR Lab, SRI International\\
	333 Ravenswood Ave\\Menlo Park, CA 94025\\
	E-mail: anand@speech.sri.com}
\runningtitle{Word discovery in continuous speech}
\runningauthor{Venkataraman}
\issue{*}{*}{*}

\newcommand{\argmax}[1]{\underset{#1}{\rm argmax}\quad}
\newcommand{\argmin}[1]{\underset{#1}{\rm argmin}\quad}

\newcommand{\W}{{\bf W}}
\newcommand{\Lex}{{\bf L}}

\newcommand{\p}{{\rm P}}
\newcommand{\C}{{\bf C}}
\newcommand{\X}{{\bf X}}

\begin{document}
\begin{singlespace}
\maketitle

\begin{abstract}
A statistical model for segmentation and word discovery in continuous
speech is presented.  An incremental unsupervised learning algorithm
to infer word boundaries based on this model is described.  Results
of empirical tests showing that the algorithm is competitive with
other models that have been used for similar tasks are also
presented.
\end{abstract}
\end{singlespace}

\section*{Introduction}

English speech lacks the acoustic analog of blank spaces that people
are accustomed to seeing between words in written text.  Discovering
words in continuous spoken speech then is an interesting problem that
has been treated at length in the literature.  The issue is 
particularly prominent in the parsing of written text in languages
that do not explicitly include spaces between words, and in the domain
of child language acquisition if we assume that children start out
with little or no knowledge of the inventory of words the language
possesses.\footnote{See, however, work in \namecite{Jusczyk:IMS97} and
\namecite{Jusczyk:DSL97} that presents strong evidence in favor of a
hypothesis that children already have a reasonably powerful and
accurate lexicon at their disposal as early as 9 months of age.}
While it is undoubtedly the case that although speech lacks explicit
demarcation of word boundaries, it nevertheless possesses significant
other cues for word discovery, it is still a matter of interest to see
exactly how much can be achieved without the incorporation of these
other cues, that is, we are interested in the performance of a {\em
bare-bones}\/ language model.  For example, there is much evidence
that stress patterns \cite{Jusczyk:PDS93,Cutler:PSI87} and
phonotactics of speech \cite{Mattys:PPE99} are of considerable aid in
word discovery.  But a bare-bones statistical model is still useful in
that it allows us to quantify precise improvements in performance upon
the integration of each specific cue into the model.  We present and
evaluate one such statistical model in this
paper.\footnote{Implementations of all the programs discussed in this
paper and the input corpus are readily available upon request from the
author.  The programs (totaling about 900 lines) have been written in
C++ to compile under Unix/Linux.  The author will assist in porting it
to other architectures or to versions of Unix other than Linux or
SunOS/Solaris if required.}

The main contributions of this study are as follows: First, it
demonstrates the applicability and competitiveness of a conservative
traditional approach for a task for which nontraditional approaches
have been proposed even recently
\cite{Brent:EPS99,Brent:DRP96,deMarcken:UAL95,Elman:FST90,Christiansen:LSS98}.
Second, although the model leads to the development of an algorithm
that learns the lexicon in an unsupervised fashion, results of partial
supervision are presented, showing that its performance is consistent
with results from learning theory.  Third, the study extends previous
work to higher-order $n$-grams, specifically up to trigrams and
discusses the results in their light.  Finally, results of experiments
suggested in Brent (1999) regarding different ways of estimating
phoneme probabilities are also reported.  Wherever possible, results
are averaged over 1000 repetitions of the experiments, thus removing
any potential advantages the algorithm may have had due to ordering
idiosyncrasies within the input corpus.

Section~\ref{sec:related} briefly discusses related literature in the
field and recent work on the same topic.  The model is described in
Section~\ref{sec:model}.  Section~\ref{sec:method} describes an
unsupervised learning algorithm based directly on the model developed
in Section~\ref{sec:model}.  This section also describes the data
corpus used to test the algorithms and the methods used.  Results are
presented and discussed in Section~\ref{sec:results}.  Finally, the
findings in this work are summarized in Section~\ref{sec:summary}.

\section{Related Work}
\label{sec:related}

While there exists a reasonable body of literature with regard to text
segmentation, especially with respect to languages such as Chinese and
Japanese, which do not explicitly include spaces between words, most
of the statistically based models and algorithms tend to fall into the
supervised learning category.  These require the model to first be
trained on a large corpus of text before it can segment its
input.\footnote{See for example \namecite{Zimin98:CTS93}.}  It is only
recently that interest in unsupervised algorithms for text
segmentation seems to have gained ground.  A notable exception in this
regard is the work by \namecite{Ando:USS99} which tries to infer word
boundaries from character $n$-gram statistics of Japanese Kanji
strings.  For example, a decision to insert a word boundary between
two characters is made solely based on whether character $n$-grams
adjacent to the proposed boundary are relatively more frequent than
character $n$-grams that straddle it.  This algorithm, however, is not
based on a formal statistcal model and is closer in spirit to
approaches based on transitional probability between phonemes or
syllables in speech.  One such approach derives from experiments by
\namecite{Saffran:WSR96} suggesting that young children might place
word boundaries between two syllables where the second syllable is
{\em surprising}\/ given the first.  This technique is described and
evaluated in \namecite{Brent:EPS99}.  Other approaches not based on
explicit probability models include those based on information
theoretic criteria such as minimum description length
\cite{Brent:DRP96,deMarcken:UAL95} and simple recurrent networks
\cite{Elman:FST90,Christiansen:LSS98}.  The maximum likelihood
approach due to \namecite{Olivier:SGL68} is probabilistic (see also
\namecite{Batchelder:CEU97}) in the sense that it is geared toward
explicitly calculating the most probable segmentation of each block of
input utterances.  However, the algorithm involves heuristic steps in
periodic purging of the lexicon and in the creation of new words in
it.  Furthermore, this approach is again not based on a formal
statistical model.  Model Based Dynamic Programming, hereafter
referred to as MBDP-1 \cite{Brent:EPS99}, is probably the most recent
work that addresses exactly the same issue as that considered in this
paper.  Both the approach presented in this paper and Brent's MBDP-1
are unsupervised approaches based on explicit probability models.  To
avoid needless repetition, we describe only Brent's MBDP-1 and direct
the interested reader to \namecite{Brent:EPS99}, which provides an
excellent review and evaluation of many of the algorithms mentioned
above.

\subsection*{Brent's model-based dynamic programming method}

\namecite{Brent:EPS99} describes a model-based approach to inferring
word boundaries in child-directed speech.  As the name implies, this
technique uses dynamic programming to infer the best segmentation.  It
is assumed that the entire input corpus, consisting of a concatenation
of all utterances in sequence, is a single event in probability space
and that the best segmentation of each utterance is implied by the
best segmentation of the corpus itself.  The model thus focuses on
explicitly calculating probabilities for every possible segmentation
of the entire corpus, subsequently picking the segmentation with the
maximum probability.  More precisely, the model attempts to calculate
$$
\p(\bar{w}_m) = \sum_n \sum_L \sum_f \sum_s
	\p(\bar{w}_m|n,L,f,s) \cdot \p(n,L,f,s)
$$
for each possible segmentation of the input corpus where the left-hand
side is the exact probability of that particular segmentation of the
corpus into words $\bar{w}_m = w_1 w_2 \cdots w_m$ and the sums are
over all possible numbers of words, $n$, in the lexicon, all possible
lexicons, $L$, all possible frequencies, $f$, of the individual words
in this lexicon and all possible orders of words, $s$, in the
segmentation.  In practice, the implementation uses an incremental
approach that computes the best segmentation of the entire corpus up
to step $i$, where the $i$th step is the corpus up to and including
the $i$th utterance.  Incremental performance is thus obtained by
computing this quantity anew after each segmentation $i-1$, assuming,
however, that segmentations of utterances up to but not including $i$
are fixed.

There are two problems with this approach.  First, the assumption
that the entire corpus of observed speech be treated as a single event
in probability space appears rather radical.  This fact is appreciated
even in \namecite[p.89]{Brent:EPS99}, which states ``{\em From a
cognitive perspective, we know that humans segment each utterance they
hear without waiting until the corpus of all utterances they will ever
hear becomes available}.''  Thus, although the incremental algorithm in
\namecite{Brent:EPS99} is consistent with a developmental model, the
formal statistical model of segmentation is not.

Second, making the assumption that the corpus is a single event in
probability space significantly increases the computational complexity
of the incremental algorithm.  The approach presented in this paper
circumvents these problems through the use of a conservative
statistical model that is directly implementable as an incremental
algorithm.  In the following section, we describe the model and how
its 2-gram and 3-gram extensions are adapted for implementation.

\section{Model Description}
\label{sec:model}

The language model described here is fairly standard in nature.  The
interested reader is referred to \namecite[p.57--78]{Jelinek:SMS97},
where a detailed exposition can be found.  Basically, we seek
\begin{eqnarray}
\label{eqn:w2}
\hat{\W} &=& \argmax{\W} \p(\W)\\
         &=& \argmax{\W} \prod_{i=1}^n \p(w_i|w_1, \cdots, w_{i-1})\\
         &=& \argmin{\W} \sum_{i=1}^n -\log \p(w_i|w_1, \cdots,
         w_{i-1})
\end{eqnarray}
where $\W = w_1, \cdots, w_n$ with $w_i \in \Lex$ denotes a particular
string of $n$ words belonging to a lexicon $\Lex$.  

The usual $n$-gram approximation is made by grouping histories $w_1,
\cdots, w_{i-1}$ into equivalence classes, allowing us to collapse
contexts into histories at most $n-1$ words backwards (for $n$-grams).
Estimations of the required $n$-gram probabilities are then done with
relative frequencies using {\em back-off} to lower-order $n$-grams
when a higher-order estimate is not reliable enough \cite{Katz:EPF87}.
Back-off is done using the \namecite{Witten:ZFP91} technique, which
allocates a probability of $N_i/(N_i+S_i)$ to unseen $i$-grams at each
stage, with the final back-off from unigrams being to an open
vocabulary where word probabilities are calculated as a normalized
product of phoneme or letter probabilities.  Here, $N_i$ is the number
of distinct $i$-grams and $S_i$ is the sum of their frequencies.  The
model can be summarized as follows:

\begin{eqnarray}
\p(w_i|w_{i-2},w_{i-1}) &=& \left\{
	\begin{array}{ll}
	\frac{S_3}{N_3 + S_3} \frac{C(w_{i-2},w_{i-1},w_i)}{C(w_{i-1},w_i)} &
	{\rm if\ } C(w_{i-2},w_{i-1},w_i) > 0 \\
	\frac{N_3}{N_3 + S_3}\p(w_i|w_{i-1}) & {\rm otherwise}\\
	\end{array}
	\right. \label{eqn:smooth3}\\
\p(w_i|w_{i-1}) &=& \left\{
	\begin{array}{ll}
	\frac{S_2}{N_2 + S_2} \frac{C(w_{i-1},w_i)}{C(w_i)} &
	{\rm if\ } C(w_{i-1},w_i) > 0 \\
	\frac{N_2}{N_2 + S_2} \p(w_i) & {\rm otherwise}\\
	\end{array}
	\right. \label{eqn:smooth2}\\
\p(w_i) &=& \left\{
	\begin{array}{ll}
	\frac{C(w_i)}{N_1 + S_1} & {\rm if\ } C(w_i) > 0 \\
	\frac{N_1}{N_1 + S_1} \p_\Sigma(w_i) & {\rm otherwise}\\
	\end{array}
	\right. \label{eqn:smooth1}\\
\p_\Sigma(w_i) &=& \frac{r(\#)\prod\limits_{j=1}^{k_i}
	r(w_i[j])}{1-r(\#)} \label{eqn:pnovel}
\end{eqnarray}
where $C()$ denotes the count or frequency function, $k_i$ denotes the
length of word $w_i$, excluding the sentinel character, `\#', $w_i[j]$
denotes its $j$th phoneme, and $r()$ denotes the relative frequency
function.  The normalization by dividing using $1-r(\#)$ in
Equation~(\ref{eqn:pnovel}) is necessary because otherwise
\begin{eqnarray}
\sum_w \p(w) &=& \sum_{i=1}^\infty (1-\p(\#))^i\p(\#)\\
&=& 1-\p(\#)
\end{eqnarray}
Since we estimate $\p(w[j])$ by $r(w[j])$, dividing by $1-r(\#)$ will ensure
that $\sum_w \p(w) = 1.$

\section{Method}
\label{sec:method}

As in \namecite{Brent:EPS99}, the model described in
Section~\ref{sec:model} is presented as an incremental learner.  The
only knowledge built into the system at start-up is the phoneme table
with a uniform distribution over all phonemes, including the sentinel
phoneme.  The learning algorithm considers each utterance in turn and
computes the most probable segmentation of the utterance using a
Viterbi search \cite{Viterbi:EBC67} implemented as a dynamic
programming algorithm described shortly.  The most likely placement of
word boundaries computed thus is committed to before considering the
next presented utterance.  Committing to a segmentation involves
learning unigram, bigram and trigram, as well as phoneme frequencies
from the inferred words.  These are used to update the respective
tables.

To account for effects that any specific ordering of input utterances
may have on the segmentations that are output, the performance of the
algorithm is averaged over 1000 runs, with each run receiving as input
a random permutation of the input corpus.

\subsection{The input corpus}

The corpus, which is identical to the one used by
\namecite{Brent:EPS99}, consists of orthographic transcripts made by
\namecite{Bernstein:PPC87} from the CHILDES collection
\cite{MACWHINNEY:CLD85}.  The speakers in this study were nine mothers
speaking freely to their children, whose ages averaged 18 months
(range 13--21).  Brent and his colleagues transcribed the corpus
phonemically (using the ASCII phonemic representation in the appendix
to this paper) ensuring that the number of subjective judgments in the
pronunciation of words was minimized by transcribing every occurrence
of the same word identically.  For example, ``look'', ``drink'' and
``doggie'' were always transcribed ``lUk'', ``drINk'' and ``dOgi''
regardless of where in the utterance they occurred and which mother
uttered them in what way.  Thus transcribed, the corpus consists of a
total of 9790 such utterances and 33,399 words including one space
after each word and one newline after each utterance.  For purposes of
illustration, Table~\ref{tbl:corpus} lists the first 20 such
utterances from a random permutation of this corpus.

It is noteworthy that the choice of this particular corpus for
experimentation is motivated purely by its use in
\namecite{Brent:EPS99}.  As has been pointed out by reviewers of an
earlier version of this paper, the algorithm is equally applicable to
plain text in English or other languages.  The main advantage of the
CHILDES corpus is that it allows for ready comparison with results
hitherto obtained and reported in the literature.  Indeed, the relative
performance of all the discussed algorithms is mostly unchanged when
tested on the 1997 Switchboard telephone speech corpus with disfluency
events removed.

\begin{singlespace}
\begin{table}
\begin{center}
\begin{tabular}{|l|l|} \hline
{\bf Phonemic Transcription} & {\bf Orthographic English text} \\ \hline \hline
hQ sIli 6v mi & How silly of me \\
lUk D*z D6 b7 wIT hIz h\&t & Look, there's the boy with his hat \\
9 TINk 9 si 6nADR bUk & I think I see another book \\
tu & Two \\
DIs wAn & This one \\
r9t WEn De wOk & Right when they walk \\
huz an D6 tEl6fon \&lIs & Who's on the telephone, Alice? \\
sIt dQn & Sit down \\
k\&n yu fid It tu D6 dOgi & Can you feed it to the doggie? \\
D* & There \\
du yu si hIm h( & Do you see him here? \\
lUk & Look \\
yu want It In & You want it in \\
W* dId It go & Where did it go? \\
\&nd WAt \# Doz & And what are those? \\
h9 m6ri & Hi Mary \\
oke Its 6 cIk & Okay it's a chick \\
y\& lUk WAt yu dId & Yeah, look what you did \\
oke & Okay \\
tek It Qt & Take it out \\ \hline
\end{tabular}
\end{center}
\caption{Twenty randomly chosen utterances from the input corpus with
their orthographic transcripts.  See the appendix for a list of the
ASCII representations of the phonemes.}
\label{tbl:corpus}
\end{table}
\end{singlespace}

\subsection{Algorithm}
\newcommand{\seg}{{\bf seg}}
\newcommand{\word}{{\bf word}}

The dynamic programming algorithm finds the most probable word
sequence for each input utterance by assigning to each segmentation a
score equal to its probability and committing to the segmentation with
the highest score.  In practice, the implementation computes the
negative logarithm of this score and thus commits to the segmentation
with the least negative logarithm of the probability.  The algorithm
for the unigram language model is presented in recursive form in
Figure~\ref{fig:dyn-rec} for readability.  The actual implementation,
however, used an iterative version.  The algorithm to evaluate the
back-off probability of a word is given in Figure~\ref{fig:word-len}.
Algorithms for bigram and trigram language models are straightforward
extensions of that given for the unigram model.  Essentially, the
algorithm description can be summed up semiformally as follows: For
each input utterance $u$, we evaluate every possible way of segmenting
it as $u = u' + w$ where $u'$ is a subutterance from the beginning of
the original utterance upto some point within it and $w$, the lexical
difference between $u$ and $u'$, is treated as a word.  The
subutterance $u'$ is itself evaluated recursively using the same
algorithm.  The base case for recursion when the algorithm rewinds is
obtained when a subutterance cannot be split further into a smaller
component subutterance and word, that is, when its length is zero.
Suppose for example, that a given utterance is {\em abcde}, where the
letters represent phonemes.  If $\seg(x)$ represents the best
segmentation of the utterance $x$ and $\word(x)$ denotes that $x$ is
treated as a word, then
$$
\seg(abcde) = {\bf best \,\, of\,\,} \left\{\begin{array}{l}
	\word(abcde) \\
	\seg(a) + \word(bcde)\\
	\seg(ab) + \word(cde)\\
	\seg(abc) + \word(de)\\
	\seg(abcd) + \word(e) \end{array} 
	\right.
$$
The {\bf evalUtterance} algorithm in Figure~\ref{fig:dyn-rec} does
precisely this.  It initially assumes the entire input utterance to be
a word on its own by assuming a single segmentation point at its right
end. It then compares the log probability of this segmentation
successively to the log probabilities of segmenting it into all
possible subutterance, word pairs.

The implementation maintains four separate tables internally, one each
for unigrams, bigrams and trigrams and one for phonemes.  When the
procedure is initially started, all the internal $n$-gram tables are
empty.  Only the phoneme table is populated with equipossible
phonemes.  As the program considers each utterance in turn and commits
to its best segmentation according to the {\bf evalUtterance}
algorithm, the various internal $n$-gram tables are updated
correspondingly.  For example, after some utterance ``abcde'' is
segmented into ``a bc de'', the unigram table is updated to increment
the frequencies of the three entries ``a'', ``bc'' and ``de'' each by
1, the bigram table to increment the frequencies of the adjacent
bigrams ``a bc'' and ``bc de'' and the trigram table to increment the
frequency of the trigram ``a bc de''.\footnote{Amending the algorithm
to include special markers for the start and end of each utterance was
not found to make a significant difference in its performance.}
Furthermore, the phoneme table is updated to increment the frequencies
of each of the phonemes in the utterance including one sentinel for
each word inferred.\footnote{In this context, see also
Section~\ref{sec:phoneme-est} regarding experiments with different
ways of estimating phoneme probabilities.}  Of course, incrementing
the frequency of a currently unknown $n$-gram is equivalent to
creating a new entry for it with frequency 1.  Note that the very
first utterance is necessarily segmented as a single word.  Since all
the $n$-gram tables are empty when the algorithm attempts to segment
it, all probabilities are necessarily computed from the level of
phonemes up.  Thus, the more words in it, the more sentinel characters
that will be included in the probability calculation and so that much
lesser will be the corresponding segmentation probability.  As the
program works its way through the corpus, correctly inferred $n$-grams
by virtue of their relatively greater preponderance compared to noise
tend to dominate their respective $n$-gram distributions and thus
dictate how future utterances are segmented.

\begin{figure}[htb]
\begin{small}
\subsubsection{Algorithm: evalUtterance}
\begin{verbatim}
BEGIN
   Input (by ref) utterance u[0..n] where u[i] are the characters in it.

   bestSegpoint := n;
   bestScore := evalWord(u[0..n]);
   for i from 0 to n-1; do
      subUtterance := copy(u[0..i]);
      word := copy(u[i+1..n]);
      score := evalUtterance(subUtterance) + evalWord(word);
      if (score < bestScore); then
         bestScore = score;
         bestSegpoint := i;
      fi
   done
   insertWordBoundary(u, bestSegpoint)
   return bestScore;
END
\end{verbatim}
\end{small}
\caption{Recursive optimization algorithm to find the best
segmentation of an input utterance using the unigram language model
described in this paper.}
\label{fig:dyn-rec}
\end{figure}

\begin{figure}[htb]
\begin{small}
\subsubsection{Function: evalWord}
\begin{verbatim}
BEGIN
   Input (by reference) word w[0..k] where w[i] are the phonemes in it.

   score = 0;
   if L.frequency(word) == 0; then {
      escape = L.size()/(L.size()+L.sumFrequencies())
      P_0 = phonemes.relativeFrequency('#');
      score = -log(escape) -log(P_0/(1-P_0));
      for each w[i]; do
         score -= log(phonemes.relativeFrequency(w[i]));
      done
   } else {
      P_w = L.frequency(w)/(L.size()+L.sumFrequencies());
      score = -log(P_w);
   }
   return score;
END
\end{verbatim}
\end{small}
\caption{The function to compute $-\log \p(w)$ of an input word $w$.
L stands for the lexicon object.  If the word is novel, then it
backs off to using a distribution over the phonemes in the word.}
\label{fig:word-len}
\end{figure}

One can easily see that the running time of the program is $O(mn^2)$
in the total number of utterances ($m$) and the length of each
utterance ($n$), assuming an efficient implementation of a hash table
allowing nearly constant lookup time is available.  Since individual
utterances typically tend to be small, especially in child-directed
speech as evidenced in Table~\ref{tbl:corpus}, the algorithm
practically approximates to a linear time procedure.  A single run
over the entire corpus typically completes in under 10 seconds on a
300 MHz i686-based PC running Linux 2.2.5-15.

Although the algorithm is presented as an unsupervised learner, a
further experiment to test the responsiveness of each algorithm to
training data is also described.  The procedure involved reserving
for training increasing amounts of the input corpus from 0\% in steps
of approximately 1\% (100 utterances).  During the training period,
the algorithm is presented the correct segmentation of the input
utterance, which it uses to update trigram, bigram, unigram and phoneme
frequencies as required.  After the initial training segment of the
input corpus has been considered, subsequent utterances are then
processed in the normal way.

\subsection{Scoring}
\label{sec:scoring}

In line with the results reported in \namecite{Brent:EPS99}, three
scores are reported --- precision, recall and lexicon precision.
Precision is defined as the proportion of predicted words that are
actually correct.  Recall is defined as the proportion of correct
words that were predicted.  Lexicon precision is defined as the
proportion of words in the predicted lexicon that are correct.  In
addition to these, the number of correct and incorrect words in the
predicted lexicon were computed, but they are not graphed here because
lexicon precision is a good indicator of both.

Precision and recall scores are computed incrementally and
cumulatively within scoring blocks, each of which consisted of 100
utterances.  These scores are computed and averaged only for the
utterances within each block scored and thus they represent the
performance of the algorithm only on the block scored, occurring in
the exact context among the other scoring blocks.  Lexicon scores
carried over blocks cumulatively.  Precision, recall and lexicon
precision scores of the algorithm in the case when it used various
amounts of training data are computed over the entire corpus.  All
scores are reported as percentages.

\section{Results}
\label{sec:results}

\begin{figure}[htb]
\begin{center}
  \includegraphics[width=11.8cm]{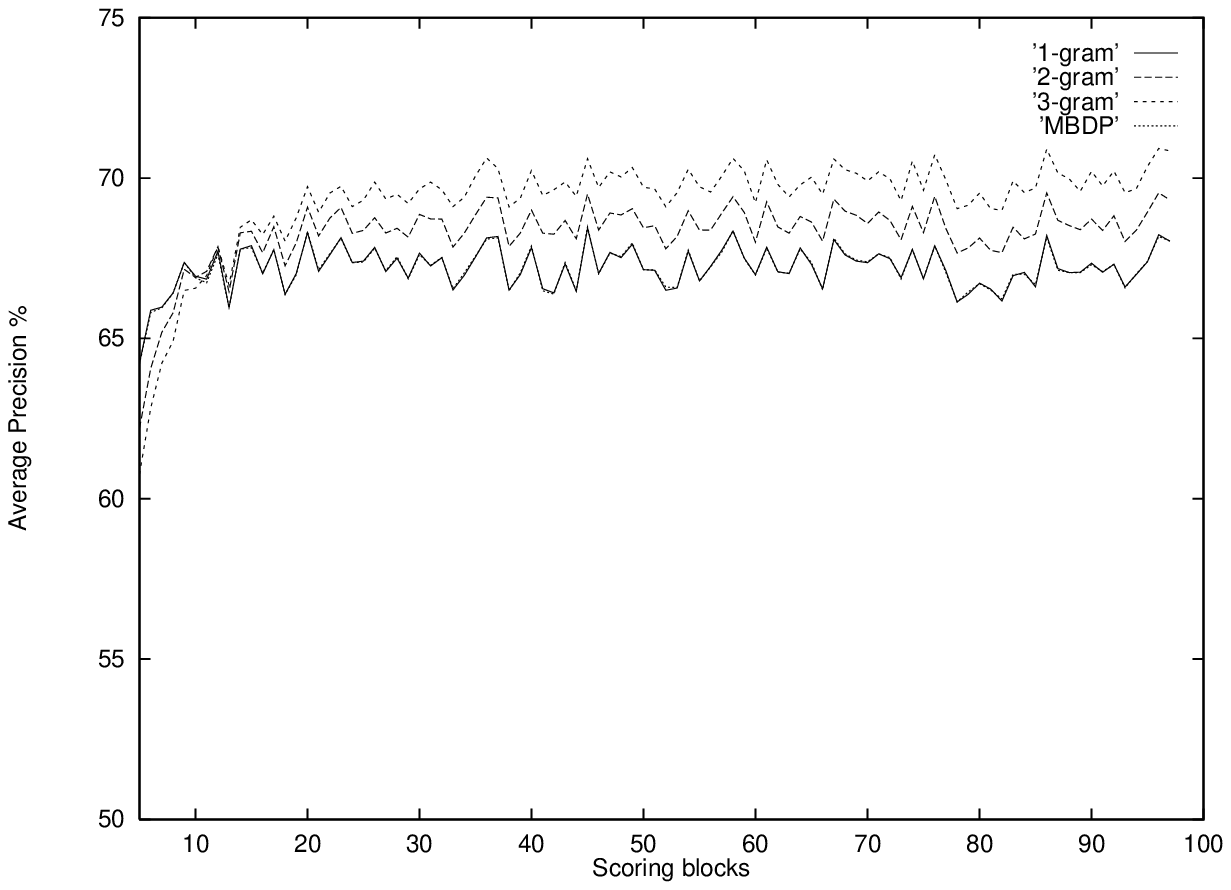}
  \caption{Averaged precision.  This is a plot of the segmentation
  precision over 100 utterance blocks averaged over 1000 runs, each
  using a random permutation of the input corpus.  Precision is
  defined as the percentage of identified words that are correct, as
  measured against the target data.  The horizontal axis represents
  the number of blocks of data scored, where each block represents 100
  utterances. The plots show the performance of the 1-gram, 2-gram,
  3-gram and MBDP-1 algorithms.  The plot for MBDP-1 is not visible
  because it coincides almost exactly with the plot for the 1-gram
  model.  Discussion of this level of similarity is provided in
  Section~\ref{sec:similarities}.  The performance of related
  algorithms due to Elman (1990), Olivier (1968) and others is
  implicitly available in this and the following graphs since Brent
  (1999) demonstrates that these all perform significantly worse than
  MBDP-1.
}  
\label{fig:r-pre}
\end{center}
\end{figure}

\begin{figure}[htb]
\begin{center}
  \includegraphics[width=11.8cm]{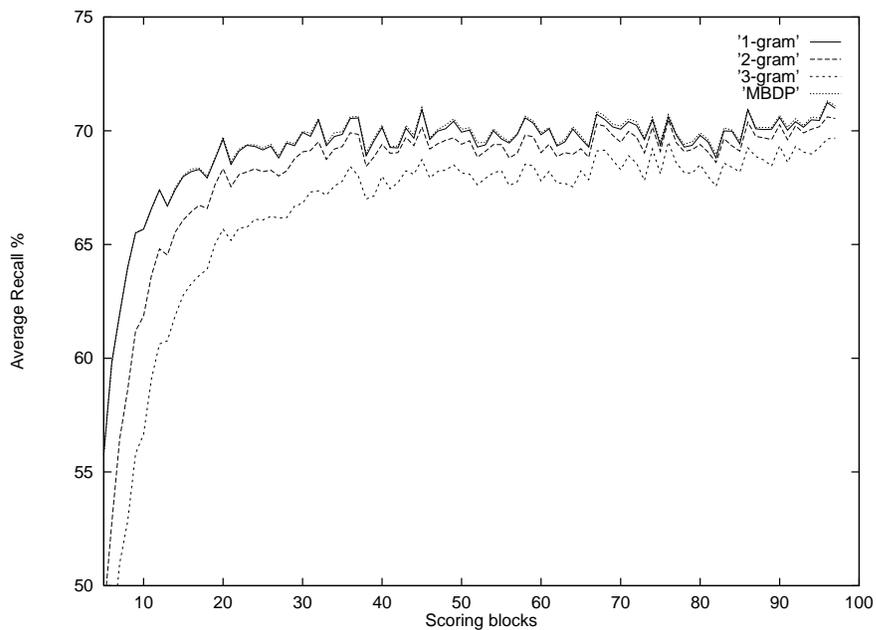}
  \caption{Averaged recall over 1000 runs, each using a random
  permutation of the input corpus.}
  \label{fig:r-rec}
\end{center}
\end{figure}

\begin{figure}[htb]
\begin{center}
  \includegraphics[width=11.8cm]{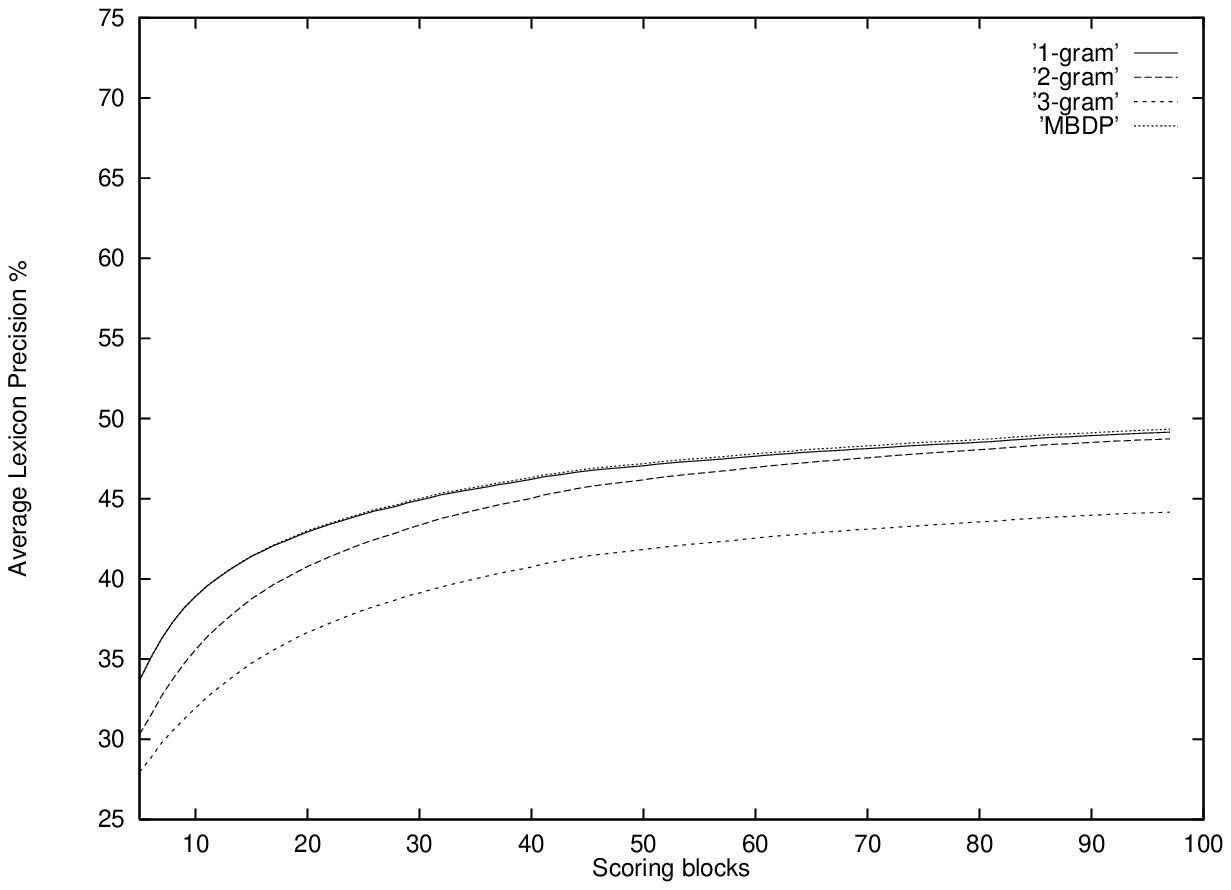}
  \caption{Averaged lexicon precision over 1000 runs, each using a
  random permutation of the input corpus.}
  \label{fig:r-lex}
\end{center}
\end{figure}

Figures~\ref{fig:r-pre}--\ref{fig:r-lex} plot the precision, recall
and lexicon precision of the proposed algorithm for each of the
unigram, bigram and trigram models against the MBDP-1 algorithm.
Although the graphs compare the performance of the algorithm with only
one published result in the field, comparison with other related
approaches is implicitly available.  \namecite{Brent:EPS99} reports
results of running the algorithms due to \namecite{Elman:FST90} and
\namecite{Olivier:SGL68}, and also algorithms based on mutual
information and transitional probability between pairs of phonemes,
over exactly the same corpus.  These are all shown to perform
significantly worse than Brent's MBDP-1.  The random baseline
algorithm in \namecite{Brent:EPS99} which consistently performs with
under 20\% precision and recall, is not graphed for the same
reason.  This baseline algorithm is given an important advantage: It
knows the exact number of word boundaries, although it does not know
their locations.  Brent argued that if MBDP-1 performs as well as this
random baseline, then at the very least, it suggests that the
algorithm is able to infer information equivalent to knowing the right
number of word boundaries.  A second important reason for not graphing
the algorithms with worse performance is that the scale on the
vertical axis could be expanded significantly by their omission, thus
allowing distinctions between the plotted graphs to be seen more
clearly.

The plots originally given in \namecite{Brent:EPS99} are over blocks
of 500 utterances.  However, because they are a result of running the
algorithm on a single corpus, there is no way of telling if the
performance of the algorithm was influenced by any particular
ordering of the utterances in the corpus.  A further undesirable
effect of reporting results of a run on exactly one ordering of the
input is that there tends to be too much variation between the values
reported for consecutive scoring blocks.  To account for both of these
problems, we report averaged results from running the algorithms on
1000 random permutations of the input data.  This has the beneficial
side effect of allowing us to plot with higher granularity since there
is much less variation in the precision and recall scores.  They are
now clustered much closer to their mean values in each block, allowing
a block size of 100 to be used to score the output.  These plots are
thus much more readable than those obtained without such averaging of
the results.

One may object that the original transcripts carefully preserve the
order of utterances directed at children by their mothers, and hence
randomly permuting the corpus would destroy the fidelity of the
simulation.  However, as we argued, the permutation and averaging does
have significant beneficial side effects, and if anything, it only
eliminates from the point of view of the algorithms the important
advantage that real children may be given by their mothers through a
specific ordering of the utterances.  In any case, we have found no
significant difference in performance between the permuted and
unpermuted cases as far as the various algorithms were concerned.  

In this context, we are curious to see how the algorithms would fare
if in fact the utterances were favorably ordered, that is, in order of
increasing length.  Clearly, this is an important advantage for all
concerned algorithms.  The results of experimenting with a
generalization of this situation, where instead of ordering the
utterances favorably, we treat an initial portion of the corpus as a
training component effectively giving the algorithms free word
boundaries after each word, are presented in
Section~\ref{sec:training}.


\subsection{Discussion}

Clearly, the performance of the present model is competitive with
MBDP-1 and as a consequence with other algorithms evaluated in
\namecite{Brent:EPS99}.  However, we note that the model proposed in
this paper has been entirely developed along conventional lines and
has not made the somewhat radical assumption of treating the entire
observed corpus as a single event in probability space.  Assuming that
the corpus consists of a single event, as Brent does, requires the
explicit calculation of the probability of the lexicon in order to
calculate the probability of any single segmentation.  This
calculation is a nontrivial task since one has to sum over all
possible orders of words in $\Lex$.  This fact is recognized in
\namecite{Brent:EPS99}, where the expression for $\p(\Lex)$ is derived
in Appendix~1 of his paper as an approximation.  One can imagine then
that it will be correspondingly more difficult to extend the language
model in \namecite{Brent:EPS99} past the case of unigrams.  As a
practical issue, recalculating lexicon probabilities before each
segmentation increases the running time of an implementation of the
algorithm.  Although all the discussed algorithms tend to complete
within one minute on the reported corpus, MBDP-1's running time is
quadratic in the number of utterances, while the language models
presented here enable computation in almost linear time.  The typical
running time of MBDP-1 on the 9790-utterance corpus averages around 40
seconds per run on a 300 MHz i686 PC while the 1-gram, 2-gram and
3-gram models average around 7, 10 and 14 seconds respectively.

Furthermore, the language models presented in this paper estimate
probabilities as relative frequencies using commonly used back-off
procedures and so they do not assume any priors over integers.
However, MBDP-1 requires the assumption of two distributions over
integers, one to pick a number for the size of the lexicon and another
to pick a frequency for each word in the lexicon.  Each is assumed
such that the probability of a given integer $\p(i)$ is given by
$\frac{6}{\pi^2i^2}$.  We have since found some evidence suggesting
that the choice of a particular prior does not have any significant
advantage over the choice of any other prior.  For example, we have
tried running MBDP-1 using $\p(i)=2^{-i}$ and still obtained
comparable results.  It is noteworthy, however, that no such
subjective prior needs to be chosen in the model presented in this
paper.

The other important difference between MBDP-1 and the present model is
that MBDP-1 assumes a uniform distribution over all possible word
orders.  That is, in a corpus that contains $n_k$ distinct words such
that the frequency in the corpus of the $i$th distinct word is given
by $f_k(i)$, the probability of any one ordering of the words in the
corpus is
$$
\frac{\prod_{i=1}^{n_k} f_k(i)!}{k!}
$$ 
because the number of unique orderings is precisely the reciprocal of
the above quantity.  Brent mentions that there may well be efficient
ways of using $n$-gram distributions in the same model.  The framework
presented in this paper is a formal statement of a model that lends
itself to such easy $n$-gram extensibility using the back-off scheme
proposed.  In fact, the results we present are direct extensions of
the unigram model into bigrams and trigrams.

In this context, an intriguing feature in the results is worth
discussing here.  We note that while with respect to precision, the
3-gram model is better than the 2-gram model which in turn is better
than the 1-gram model, with respect to recall their performance is
exactly the opposite.  We may attempt to explain this behavior thus:
Since the 3-gram model places greatest emphasis on word triples, which
are relatively less frequent, it has the least evidence of all from
the observed data to infer word boundaries.  Even though back-off is
performed for bigrams when a trigram is not found, there is a cost
associated with such backing off --- this is the extra fractional
factor $N_3/(N_3+S_3)$ in the calculation of the segmentation's
probability.  Consequently, the 3-gram model is the most conservative
in its predictions.  When it does infer a word boundary it is likely
to be correct.  This contributes to its relatively higher precision
since precision is a measure of the proportion of inferred boundaries
that were correct.  More often than not, however, when the 3-gram
model does not have enough evidence to infer words, it simply outputs
the default segmentation, which is a single word (the entire
utterance) instead of more than one incorrectly inferred ones.  This
contributes to its poorer recall since recall is an indicator of the
number of words the model fails to infer.  Poorer lexicon precision is
likewise explained.  Because the 3-gram model is more conservative, it
infers new words only when there is strong evidence for them.  As a
result many utterances are inserted as whole words into its lexicon
thereby contributing to decreased lexicon precision.  The presented
framework thus provides for a systematic way of trading off precision
for recall or vice-versa.  Models utilizing higher-order $n$-grams
give better recall at the expense of precision.

\subsection{Estimation of phoneme probabilities}
\label{sec:phoneme-est}

\namecite[p.101]{Brent:EPS99} suggests that it could be worthwhile
studying whether learning phoneme probabilities from distinct lexical
entries yields better results than learning these probabilities from
the input corpus.  That is, the probability of the phoneme ``th'' in
``the'' is better not inflated by the preponderance of {\em the}\/ and
{\em the}-like words in actual speech, but rather controlled by the
number of such distinct words.  We report some initial analysis and
experimental results in this regard.  Assume the existence of some
function $\Psi_X:{\bf N}\rightarrow{\bf N}$ that maps the size, $n$,
of a corpus $\C$, onto the size of some subset $\X$ of $\C$ we may
define.  If this subset $\X=\C$, then $\Psi_\C$ is the identity
function and if $\X=\Lex$ is the set of distinct words in $\C$ we have
$\Psi_\Lex(n) = |\Lex|$.

\begin{figure}[htb]
\begin{center}
  \includegraphics[width=11.8cm]{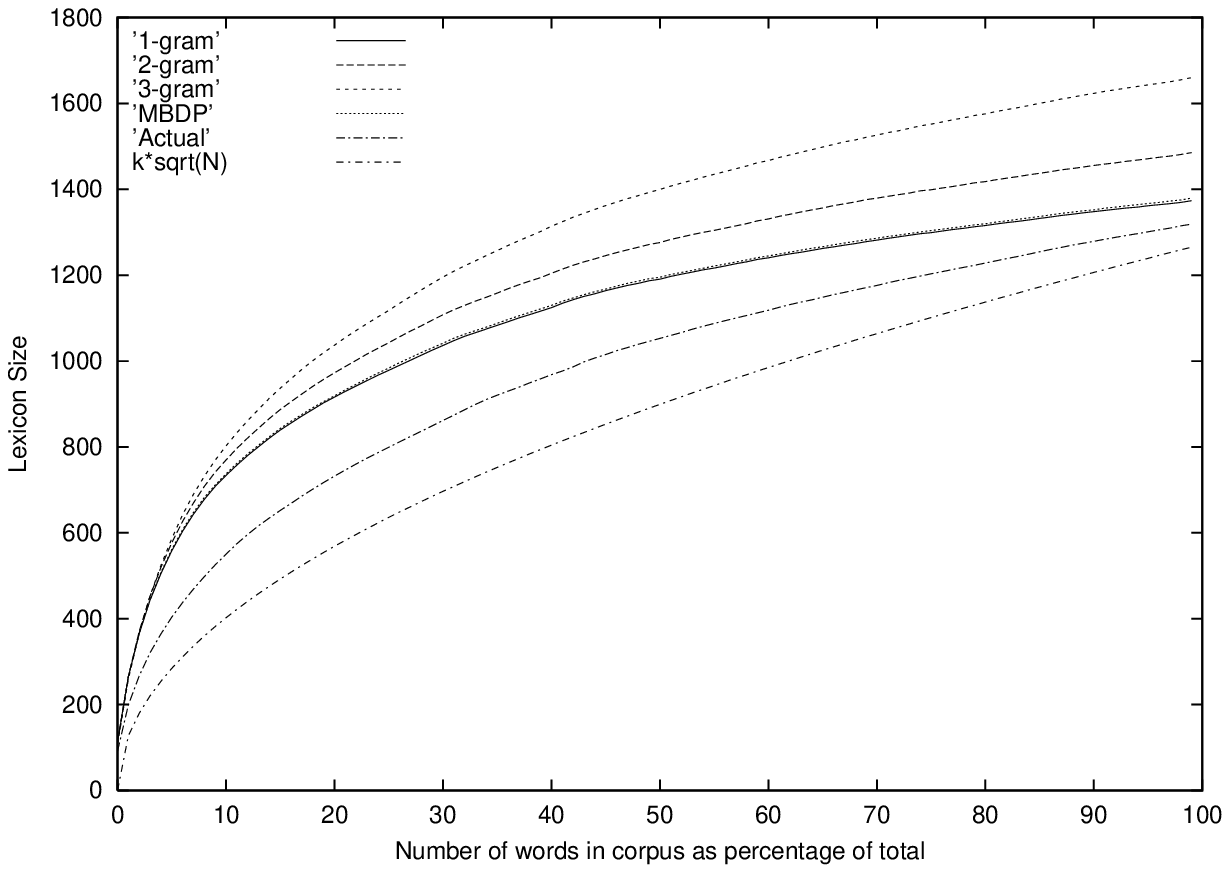} 
  \caption{Plot shows the rate of growth of the lexicon with
  increasing corpus size as percentage of total size.  {\em Actual}\/
  is the actual number of distinct words in the input corpus.  {\em
  1-gram, 2-gram, 3-gram}\/ and {\em MBDP}\/ plot the size of the
  lexicon as inferred by each of the algorithms.  It is interesting
  that the rates of lexicon growth are roughly similar to each other
  regardless of the algorithm used to infer words and that they may
  all potentially be modeled by a function such as $k\sqrt{N}$ where
  $N$ is the corpus size.}
  \label{fig:lexrate}
\end{center}
\end{figure}

Let $l_\X$ be the average number of phonemes per word in $\X$ and let
${E_a}_\X$ be the average number of occurrences of phoneme $a$ per
word in $\X$.  Then we may estimate the probability of an arbitrary
phoneme $a$ from $\X$ as follows.
\begin{eqnarray*}
\p(a|\X) &=& \frac{C(a|\X)}{\sum_{a_i} C(a_i|\X)} \\
         &=& \frac{{E_a}_\X \Psi_\X(N)}{l_\X\Psi_\X(N)}
\end{eqnarray*}
where, as before, $C(a|\X)$ is the count function that gives the
frequency of phoneme $a$ in $\X$.  If $\Psi_X$ is deterministic, we
can then write
\begin{equation}
\p(a|\X) = \frac{{E_a}_\X}{l_\X} \label{eqn:phon-Xprob}
\end{equation}
Our experiments suggest that ${E_a}_\Lex \sim {E_a}_\C$ and that
$l_\Lex \sim l_\C$.  We are thus led to suspect that estimates should
roughly be the same regardless of whether probabilities are estimated
from $\Lex$ or $\C$.  This is indeed borne out by the results we
present below.  Of course, this is true only if there exists some
deterministic function $\Psi_\Lex$ as we assumed, and this may not
necessarily be the case.  There is, however, some evidence that the
number of distinct words in a corpus can be related to the total number
of words in the corpus in this way.  In Figure~\ref{fig:lexrate} the
rate of lexicon growth is plotted against the proportion of the corpus
size considered.  The values for lexicon size were collected using the
Unix filter
\begin{quote}
{\tt cat \$*|tr ' ' $\backslash\backslash$012|awk '}\{{\tt print
(L[\$0]++)? v : ++v;}\}'
\end{quote}
and smoothed by averaging over 100 runs each on a separate permutation
of the input corpus.  That the lexicon size can be approximated by a
deterministic function of the corpus size is strongly suggested by the
the plot.  It is interesting that the shape of the plot is roughly the
same regardless of the algorithm used to infer words suggesting that
they all segment {\em word-like}\/ units that share at least some
statistical properties with actual words.

Table~\ref{tbl:phon-learn} summarizes our empirical findings in this
regard.  For each model --- namely, 1-gram, 2-gram, 3-gram and MBDP-1
--- we test all three of the following possibilities:
\begin{enumerate}
\item \, Always use a uniform distribution over phonemes.
\item \, Learn the phoneme distribution from the lexicon.
\item \, Learn the phoneme distribution from the corpus, that is, from
         all words, whether distinct or not.
\end{enumerate}

\begin{singlespace}
\begin{table}
\begin{center}
\begin{tabular}{l|l|l|l|l|}
\multicolumn{5}{c}{} \\
\multicolumn{1}{c}{}   & \multicolumn{4}{c}{\bf Precision} \\ 
                                                   \cline{2-5}
\multicolumn{1}{c|}{}  & 1-gram  & 2-gram  & 3-gram  & MBDP      \\ 
                                                   \cline{2-5}
{\bf Lexicon}          & 67.7  & 68.08 & 68.02 & 67        \\
{\bf Corpus}           & 66.25 & 66.68 & 68.2  & 66.46     \\
{\bf Uniform}          & 58.08 & 64.38 & 65.64 & 57.15     \\
\cline{2-5}\cline{2-5}
\multicolumn{5}{c}{}\\
\multicolumn{1}{c}{}   & \multicolumn{4}{c}{\bf Recall }   \\ 
                                                   \cline{2-5}
\multicolumn{1}{c|}{}  & 1-gram  & 2-gram  & 3-gram  & MBDP      \\
                                                   \cline{2-5}
{\bf Lexicon}          & 70.18 & 68.56 & 65.07 & 69.39     \\
{\bf Corpus}           & 69.33 & 68.02 & 66.06 & 69.5      \\
{\bf Uniform}          & 65.6  & 69.17 & 67.23 & 65.07     \\
\cline{2-5}\cline{2-5}
\multicolumn{5}{c}{}\\
\multicolumn{1}{c}{}   & \multicolumn{4}{c}{\bf Lexicon Precision}   \\
                                                   \cline{2-5}
\multicolumn{1}{c|}{}  & 1-gram  & 2-gram  & 3-gram  & MBDP      \\
                                                   \cline{2-5}
{\bf Lexicon}          & 52.85 & 54.45 & 47.32 & 53.56     \\
{\bf Corpus}           & 52.1  & 54.96 & 49.64 & 52.36     \\
{\bf Uniform}          & 41.46 & 52.82 & 50.8  & 40.89     \\
\cline{2-5}\cline{2-5}
\end{tabular}
\caption{Summary of results from each of the algorithms for each of
the following cases:  Lexicon -- Phoneme probabilities estimated from
the lexicon, Corpus -- Phoneme probabilities estimated from input
corpus and Uniform -- Phoneme probabilities assumed uniform and
constant.}
\label{tbl:phon-learn}
\end{center}
\end{table}
\end{singlespace}

The row labeled {\em Lexicon}\/ lists scores on the entire corpus from
a program that learned phoneme probabilities from the lexicon.  The
row labeled {\em Corpus}\/ lists scores from a program that learned
these probabilities from the input corpus, and the row labeled {\em
Uniform}\/ lists scores from a program that just assumed uniform
phoneme probabilities throughout.  While the performance is clearly
seen to suffer when a uniform distribution over phonemes is assumed,
whether the distribution is estimated from the lexicon or the corpus
does not seem to make any significant difference.  These results lead
us to believe that from an empirical point of view it really does not
matter whether phoneme probabilities are estimated from the corpus or
the lexicon.  Intuitively, however, it seems that the right approach
ought to be one that estimates phoneme frequencies from the corpus
data since frequent words ought to have a greater influence on the
phoneme distribution than infrequent ones.

\subsection{Responsiveness to training}
\label{sec:training}

\begin{figure}[htb]
\begin{center}
  \includegraphics[width=11.8cm]{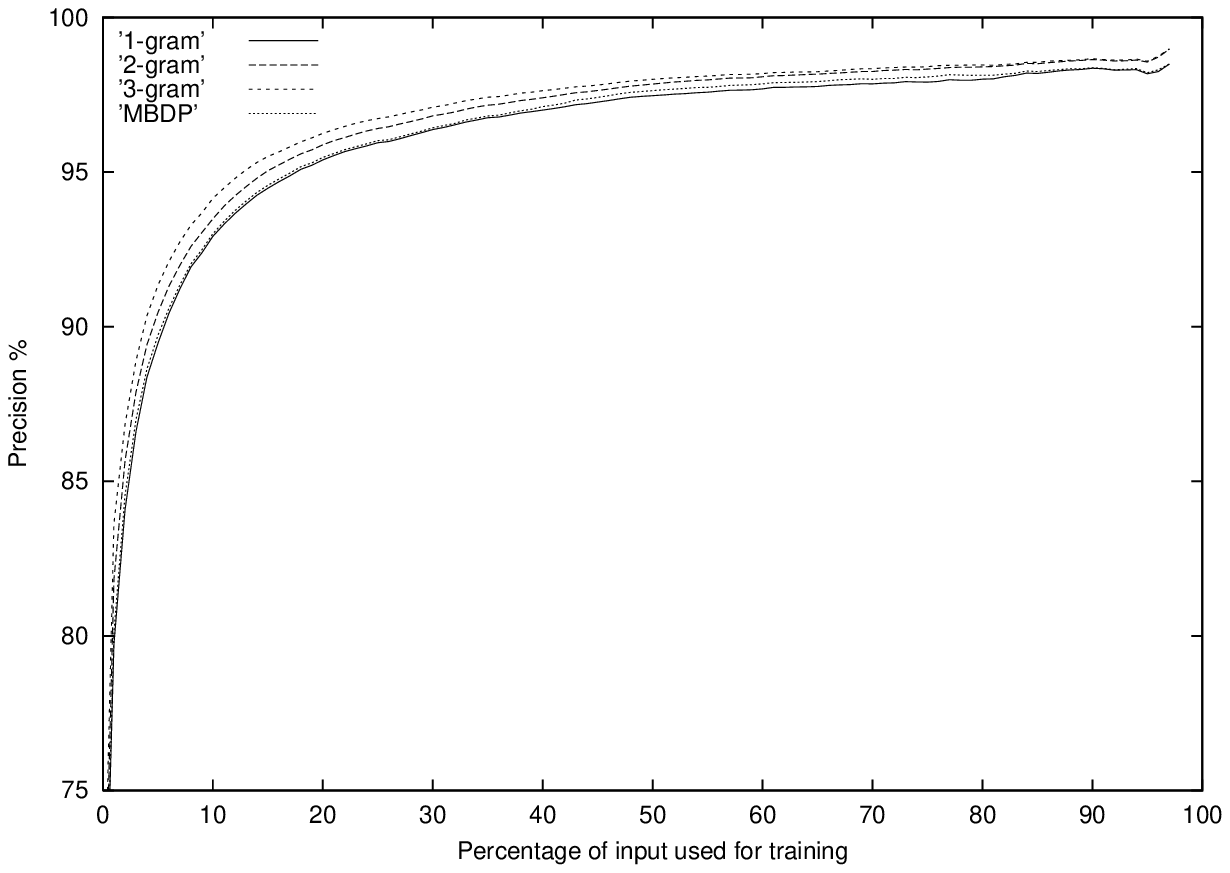}
  \caption{Responsiveness of the algorithm to training information.
  The horizontal axis represents the initial percentage of the data
  corpus that was used for training the algorithm.  This graph shows
  the improvement in precision with training size.}
  \label{fig:t-pre}
\end{center}
\end{figure}

\begin{figure}[htb]
\begin{center}
  \includegraphics[width=11.8cm]{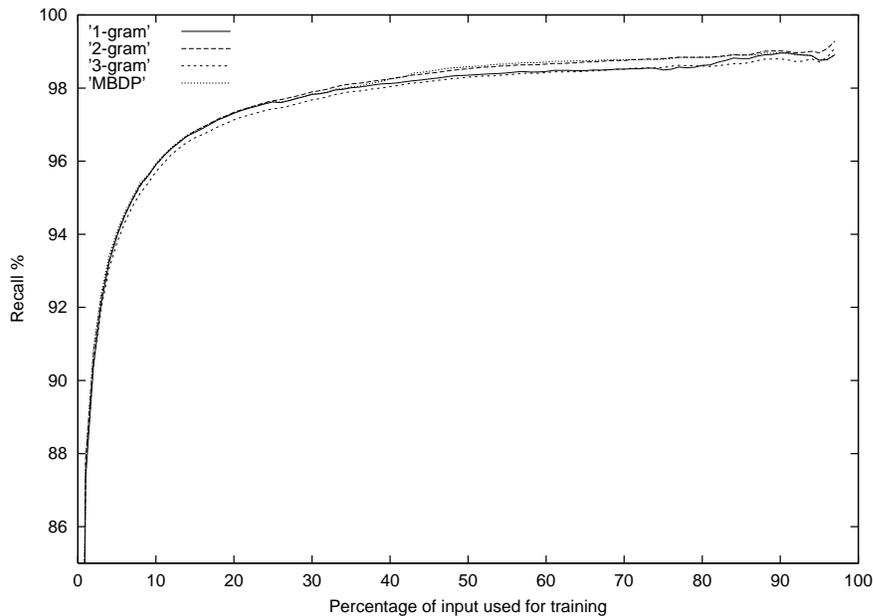}
  \caption{Improvement in recall with training size.}
  \label{fig:t-rec}
\end{center}
\end{figure}

It is interesting to compare the responsiveness of the various algorithms
to the effect of training data.  Figures~\ref{fig:t-pre}--\ref{fig:t-rec}
plot the results (precision and recall) over the whole input corpus, that
is, blocksize = $\infty$, as a function of the initial proportion of the
corpus reserved for training.  This is done by dividing the corpus into two
segments, with an initial training segment being used by the algorithm to
learn word, bigram, trigram and phoneme probabilities and the latter
actually being used as the test data.  A consequence of this is that the
amount of data available for testing becomes progressively smaller as the
percentage reserved for training grows.  So the significance of the test
would diminish correspondingly.  We may assume that the plots cease to be
meaningful and interpretable when more than about 75\% (about 7500
utterances) of the corpus is used for training.  At 0\%, there is no
training information for any algorithm and the scores are identical to
those reported earlier.  We increase the amount of training data in steps
of approximately 1\% (100 utterances).  For each training set size, the
results reported are averaged over 25 runs of the experiment, each over a
separate random permutation of the corpus.  The motivation, as before, was
both to account for ordering idiosyncrasies as well as to smooth the graphs
to make them easier to interpret.


We interpret Figures~\ref{fig:t-pre} and~\ref{fig:t-rec} as suggesting
that the performance of all discussed algorithms can be boosted
significantly with even a small amount of training.  It is
noteworthy and reassuring to see that, as one would expect from
results in computational learning theory \cite{Haussler:QIB88}, the
number of training examples required to obtain a desired value of
precision, $p$, appears to grow with $1/(1-p)$.  The intriguing
reversal in the performance of the various $n$-gram models with
respect to precision and recall is again seen here and the explanation
for this too is the same as before.  We further note, however, that
the difference in performance between the different models tends to
narrow with increasing training size, that is, as the amount of
evidence available to infer word boundaries increases, the 3-gram
model rapidly catches up with the others in recall and lexicon
precision.  It is likely, therefore, that with adequate training data,
the 3-gram model might be the most suitable one to use.  The following
experiment lends some substance to this suspicion.

\subsection{{\em Fully trained}\/ algorithms}

The preceding discussion makes us curious to see what would happen if
the above scenario was extended to the limit, that is, if 100\% of the
corpus was used for training.  This precise situation was in fact
tested.  The entire corpus was concatenated onto itself and the models
then trained on exactly the former half and tested on the latter half
of the corpus augmented thus.  Although the unorthodox nature of this
procedure requires us to not attach much significance to the outcome,
we nevertheless find the results interesting enough to warrant some
mention, and we discuss here the performance of each of the four
algorithms on the test segment of the input corpus (the latter half).
As one would expect from the results of the preceding experiments, the
trigram language model outperforms all others.  It has a precision and
recall of 100\% on the test input, except for exactly four utterances.
These four utterances are shown in Table~\ref{tbl:3gram-out}
retranscribed into plain English.
\begin{table}
\begin{center}
\begin{tabular}{|l|l|l|}
\hline
{\bf \#} & {\bf 3-gram output} & {\bf Target} \\
\hline \hline
3482 & $\cdots$ in the {\bf doghouse}           & $\cdots$ in the dog house\\
5572 & {\bf aclock}                             & a clock\\
5836 & that's {\bf alright}                     & that's all right\\
7602 & that's right it's a {\bf hairbrush}      & that's right it's a hair brush\\
\hline
\end{tabular}
\caption{Errors in the output of a fully trained 3-gram language
model.  Erroneous segmentations are shown in boldface.}
\label{tbl:3gram-out}
\end{center}
\end{table}

Intrigued as to why these errors occurred, we examined the corpus,
only to find erroneous transcriptions in the input.  ``dog house'' is
transcribed as a single word ``dOghQs'' in utterance 614, and as two
words elsewhere.  Likewise, ``o'clock'' is transcribed ``6klAk'' in
utterance 5917, ``alright'' is transcribed ``Olr9t'' in utterance 3937,
and ``hair brush'' is transcribed ``h*brAS'' in utterances 4838 and
7037.  Elsewhere in the corpus, these are transcribed as two words.

\begin{table}
\begin{center}
\begin{tabular}{|l|l|l|}
\hline
{\bf \#} & {\bf 2-gram output} & {\bf Target} \\
\hline \hline
 614 & you want the {\bf dog house}       & you want the doghouse\\
3937 & thats {\bf all right}              & that's alright\\
5572 & {\bf a clock}                      & a clock\\
7327 & look a {\bf hairbrush}             & look a hair brush\\
7602 & that's right its a {\bf hairbrush} & that's right its a hair brush\\
7681 & {\bf hairbrush}                    & hair brush\\
7849 & it's called a {\bf hairbrush}      & it's  called a hair brush\\
7853 & {\bf hairbrush}                    & hair brush\\
\hline
\end{tabular}
\caption{Errors in the output of a fully trained 2-gram language
model.  Erroneous segmentations are shown in boldface.}
\label{tbl:2gram-out}
\end{center}
\end{table}

The erroneous segmentations in the output of the 2-gram language model
are likewise shown in Table~\ref{tbl:2gram-out}.  As expected,
the effect of reduced history is apparent through an increase in the
total number of errors.  However, it is interesting to note that while
the 3-gram model incorrectly segmented an incorrect transcription
(utterance 5836) ``that's all right'' to produce ``that's alright'', the
2-gram model incorrectly segmented a correct transcription (utterance
3937) ``that's alright'' to produce ``that's all right''.  The reason for
this is that the bigram ``that's all'' is encountered relatively
frequently in the corpus and this biases the algorithm toward
segmenting the ``all'' out of ``alright'' when it follows ``that's''.
However, the 3-gram model is not likewise biased because having
encountered the exact 3-gram ``that's all right'' earlier, there is no
back-off to try bigrams at this stage.

Similarly, it is interesting that while the 3-gram model incorrectly
segments the incorrectly transcribed ``dog house'' into ``doghouse'' in
utterance 3482, the 2-gram model incorrectly segments the correctly
transcribed ``doghouse'' into ``dog house'' in utterance 614.  In the
trigram model, $-\log \p({\rm house|the,dog}) = 4.8$ and $-\log \p({\rm
dog|in, the}) = 5.4$, giving a score of 10.2 to the segmentation ``dog
house''.  However, due to the error in transcription, the trigram ``in the
doghouse'' is never encountered in the training data although the bigram
``the doghouse'' is.  Backing off to bigrams, $-\log \p({\rm
doghouse|the})$ is calculated as 8.1.  Hence the probability that
``doghouse'' is segmented as ``dog house'' is less than the probability
that it is a word by itself.  In the 2-gram model, however, $-\log \p({\rm
dog|the})\p({\rm house|dog}) = 3.7 + 3.2 = 6.9$ while $-\log \p({\rm
doghouse|the}) = 7.5$, whence ``dog house'' is the preferred segmentation
although the training data contained instances of all three bigrams.

\begin{singlespace}
\begin{table}
\begin{center}
\begin{tabular}{|l|l|l|}
\hline
{\bf \#} & {\bf 1-gram output} & {\bf Target} \\
\hline \hline
 244 & brush {\bf Alice 's} hair                 & brush Alice's hair\\
 503 & you're {\bf in to} distraction $\cdots$   & you're into distraction $\cdots$\\
1066 & you {\bf my trip} it                      & you might rip it\\
1231 & this is little {\bf doghouse}             & this is little dog house\\
1792 & stick it {\bf on to} there                & stick it ontu there\\
3056 & $\cdots$ so he doesn't run {\bf in to}    & $\cdots$ so hi doesn't run into\\
3094 & $\cdots$ to be in the {\bf highchair}     & $\cdots$ to be in the high chair\\
3098 & $\cdots$ for this {\bf highchair}         & $\cdots$ for this high chair\\
3125 & $\cdots$ {\bf already} $\cdots$           & $\cdots$ all ready $\cdots$ \\
3212 & $\cdots$ could talk {\bf in to} it        & $\cdots$ could talk into it\\
3230 & can {\bf heel I} down on them             & can he lie down on them\\
3476 & that's a {\bf doghouse}                   & that's a dog house \\
3482 & $\cdots$ in the {\bf doghouse}            & $\cdots$ in the dog house\\
3923 & $\cdots$ when {\bf it's nose}             & $\cdots$ when it snows\\
3937 & that's {\bf all right}                    & that's alright\\
4484 & its about {\bf mealtime s}                & its about meal times\\
5328 & tell him to {\bf way cup}                 & tell him to wake up\\
5572 & {\bf o'clock}                             & a clock\\
5671 & where's my little {\bf hairbrush}         & where's my little hair brush\\
6315 & that's {\bf a nye}                        & that's an i\\
6968 & okay mommy {\bf take seat}                & okay mommy takes it\\
7327 & look a {\bf hairbrush}                    & look a hair brush\\
7602 & that's right its a {\bf hairbrush}        & that's right its a hair brush\\
7607 & go {\bf along} way to find it today       & go a long way to find it today\\
7676 & mom {\bf put sit}                         & mom puts it\\
7681 & {\bf hairbrush}                           & hair brush\\
7849 & its called a {\bf hairbrush}              & its called a hair brush\\
7853 & {\bf hairbrush}                           & hair brush\\
8990 & $\cdots$ in the {\bf highchair}           & $\cdots$ in the high chair\\
8994 & for baby's a nice {\bf highchair}         & for baby's a nice high chair\\
8995 & that's like a {\bf highchair} that's right&that's like a high chair that's right\\
9168 & he has {\bf along} tongue                 & he has a long tongue\\
9567 & you wanna go in the {\bf highchair}       & you wanna go in the high chair\\
9594 & {\bf along} red tongue                    & a long red tongue\\
9674 & {\bf doghouse}                            & dog house\\
9688 & {\bf highchair} again                     & high chair again\\
9689 & $\cdots$ the {\bf highchari}              & $\cdots$ the high chair\\
9708 & I have {\bf along} tongue                 & I have a long tongue\\
\hline
\end{tabular}
\caption{Errors in the output of a fully trained 1-gram language model.}
\label{tbl:1gram-out}
\end{center}
\end{table}
\end{singlespace}

The errors in the output of a 1-gram model are shown in
Table~\ref{tbl:1gram-out}, but we do not discuss these as we did 
for the 3-gram and 2-gram outputs.  The errors in the output of
Brent's fully trained MBDP-1 algorithm are not shown here because they
are identical to those produced by the 1-gram model except for one
utterance.  This single difference is the segmentation of utterance
8999, ``lItL QtlEts'' (little outlets), which the 1-gram model
segmented correctly as ``lItL QtlEts'', but MBDP-1 segmented as ``lItL
{\bf Qt lEts}''.  In both MBDP-1 and the 1-gram model, all four words,
``little'', ``out'', ``lets'' and ``outlets'', are familiar at the time
of segmenting this utterance.  MBDP-1 assigns a score of $5.3 +
5.95 = 11.25$ to the segmentation ``out + lets'' versus a score
of $11.76$ to ``outlets''.  As a consequence, ``out + lets'' is the
preferred segmentation.  In the 1-gram language model, the
segmentation ``out + lets'' scores $5.31 +  5.97 = 11.28$,
whereas ``outlets'' scores $11.09$.  Consequently, it selects
``outlets'' as the preferred segmentation.  The only thing we could
surmise from this was either that this difference must have come about
due to chance (meaning that this may well have not been the case if
certain parts of the corpus had been any different) or else the
interplay between the different elements in the two models is too
subtle to be addressed within the scope of this paper.

\subsection{Similarities between MBDP-1 and the 1-gram Model}
\label{sec:similarities}

The similarities between the outputs of MBDP-1 and the 1-gram model
are so great that we suspect they may essentially be capturing the
same nuances of the domain.  Although \namecite{Brent:EPS99}
explicitly states that probabilities are not estimated for words, it
turns out that considering the entire corpus does end up having the
same effect as estimating probabilities from relative frequencies as
the 1-gram model does.  The {\em relative probability}\/ of a familiar
word is given in Equation~22 of \namecite{Brent:EPS99} as
$$
\frac{f_k(\hat{k})}{k}\cdot
\left( \frac{f_k(\hat{k})-1}{f_k(\hat{k})}\right)^2
$$
where $k$ is the total number of words and $f_k(\hat{k})$ is the
frequency at that point in segmentation of the $\hat{k}$th word. It 
effectively approximates to the relative frequency 
$$
\frac{f_k(\hat{k})}{k}
$$
as $f_k(\hat{k})$ grows.  The 1-gram language model of this paper
explicitly claims to use this specific estimator for the unigram
probabilities.  From this perspective, both MBDP-1 and the 1-gram
model tend to favor the segmenting out of familiar words that do not
overlap.  It is interesting, however, to see exactly how much evidence
each needs before such segmentation is carried out.  In this context,
the author recalls an anecdote recounted by a British colleague who,
while visiting the USA, noted that the populace in the vicinity of his
institution had grown up thinking that ``Damn British'' was a single
word, by virtue of the fact that they had never heard the latter word
in isolation.  We test this particular scenario here with both
algorithms.  The programs are first presented with the utterance
``damnbritish''.  Having no evidence to infer otherwise, both programs
assume that ``damnbritish'' is a single word and update their lexicons
accordingly.  The interesting question now is exactly how many
instances of the word ``british'' in isolation should either program
see before being able to successfully segment a subsequent
presentation of ``damnbritish'' correctly.

Obviously, if the word ``damn'' is also unfamiliar, there will never
be enough evidence to segment it out in favor of the familiar word
``damnbritish''.  Hence each program is presented next with two
identical utterances, ``damn''.  We do need to present two such
utterances.  Otherwise the estimated probabilities of the familiar
words ``damn'' and ``damnbritish'' will be equal.  Consequently, the
probability of any segmentation of ``damnbritish'' that contains the
word ``damn'' will be less than the probability of ``damnbritish''
considered as a single word.

At this stage, we present each program with increasing numbers of
utterances consisting solely of the word ``british'' followed by a
repetition of the very first utterance -- ``damnbritish''.  We find
that MBDP-1 needs to see the word ``british'' on its own three times
before having enough evidence to disabuse itself of the notion that
``damnbritish'' is a single word.  In comparison, the 1-gram model is
more skeptical.  It needs to see the word ``british'' on its own seven
times before committing to the right segmentation.  To illustrate the
inherent simplicity of the presented model, we can show that it is
easy to predict this number analytically from the presented 1-gram
model.  Let $x$ be the number of instances of ``british'' required.
Then using the discounting scheme described, we have
\begin{eqnarray*}
\p({\rm damnbritish}) &=& 1/(x+6)\\
\p({\rm damn}) &=& 2/(x+6)\qquad {\rm  and}\\
\p({\rm british}) &=& x/(x+6)
\end{eqnarray*}
We seek an $x$ for which $\p({\rm damn}) \p({\rm british}) > \p({\rm
damnbritish})$.  Thus, we get
$$
2x/(x+6)^2 > 1/(x+6) \Rightarrow x > 6
$$
The actual scores for MBDP-1 when presented with ``damnbritish'' for a
second time are $-\log \p({\rm damnbritish}) = 2.8$ and $-\log
\p({\rm D\&m}) -\log \p({\rm brItIS}) = 1.8 + 0.9 = 2.7$.
For the 1-gram model, $-\log \p({\rm damnbritish}) = 2.6$ and
$-\log \p({\rm D\&m}) -\log \p({\rm brItIS}) = 1.9 + 0.6 =
2.5$.  Note, however, that skepticism in this regard is not always
a bad attribute.  It helps to be skeptical in inferring new words
because a badly inferred word will adversely influence future
segmentation accuracy.

\section{Summary}
\label{sec:summary}

In summary, we have presented a formal model of word discovery in
continuous speech.  The main advantages of this model over that of
\namecite{Brent:EPS99} are, first, that the present model has been
developed entirely by direct application of standard techniques and
procedures in speech processing.  It makes few assumptions about
the nature of the domain and remains as far as possible conservative
in its development.  Finally, the model is easily extensible to
incorporate more historical detail.  This is clearly evidenced by the
extension of the unigram model to handle bigrams and trigrams.
Empirical results from experiments suggest that the algorithm performs
competitively with alternative unsupervised algorithms proposed for
inferring words from continous speech.  We have also carried out and
reported results from experiments to determine whether particular ways
of estimating phoneme (or letter) probabilities may be more suitable
than others.

Although the algorithm is originally presented as an unsupervised
learner, we have shown the effect that training data has on its
performance.  It appears that the 3-gram model is the most responsive
to training information with regard to segmentation precision,
obviously by virtue of the fact that it {\em keeps\/} more knowledge
from the presented utterances.  Indeed, we see that a {\em fully
trained}\/ 3-gram model performs with 100\% accuracy on the test set.
Admittedly, the test set in this case was identical to the training
set, but we should keep in mind that we were only keeping limited
history --- namely 3-grams --- and a significant number of utterances
in the input corpus (4023 utterances) were 4 words or more in length.
Thus, it is not completely insignificant that the algorithm was able
to perform this well.

\subsection*{Future work}

We are presently working on the incorporation of more complex phoneme
distributions into the model.  These are, namely, the biphone and triphone
models.  Some preliminary results we have obtained in this regard appear to
be encouraging.

With regard to estimation of word probabilities, modification of the
model to address the sparse data problem using interpolation such that
$$
\p(w_i|w_{i-2},w_{i-1}) = \lambda_3 f(w_i|w_{i-2},w_{i-1}) + \lambda_2
	f(w_i|w_{i-1}) + \lambda_1 f(w_i)
$$
where the positive coefficients satisfy $\lambda_1 + \lambda_2 +
\lambda_3 = 1$ and can be derived so as to maximize $\p(\W)$ is being
explored as a fruitful avenue.

Using the lead from \namecite{Brent:EPS99}, attempts to model more complex
distributions for unigrams such as those based on {\em template grammars\/}
and the systematic incorporation of prosodic, stress and phonotactic
constraint information into the model are also the subject of current
interest.  We already have some unpublished results suggesting that biasing
the segmentation such that that every word must have at least one vowel in
it dramatically increases segmentation precision from 67.7\% to 81.8\% and
imposing a constraint that words can begin or end only with permitted
clusters of consonants increases precision to 80.65\%.  We are planning
experiments to investigate models in which these properties can be learned
in the same way as $n$-grams.

\newpage
\section*{Appendix - Inventory of Phonemes}

The following tables list the ASCII representations of the phonemes
used to transcribe the corpus into a form suitable for processing by
the algorithms.
\vspace{0.5cm}

\begin{singlespace}
\begin{tabular}{|c|l|} 
\multicolumn{2}{c}{\bf Consonants}\\ \hline
{\bf ASCII} & {\bf Example}  \\ \hline \hline
p  & {\bf p}an \\
b  & {\bf b}an \\
m  & {\bf m}an \\
t  & {\bf t}an \\
d  & {\bf d}am \\
n  & {\bf n}ap \\
k  & {\bf c}an \\
g  & {\bf g}o \\
N  & si{\bf ng} \\
f  & {\bf f}an \\
v  & {\bf v}an \\
T  & {\bf th}in \\
D  & {\bf th}an \\
s  & {\bf s}and \\
z  & {\bf z}ap \\
S  & {\bf sh}ip \\
Z  & plea{\bf s}ure \\
h  & {\bf h}at \\
c  & {\bf ch}ip \\
G  & {\bf g}el \\
l  & {\bf l}ap \\
r  & {\bf r}ap \\
y  & {\bf y}et \\
w  & {\bf w}all \\
W  & {\bf wh}en \\
L  & bott{\bf le} \\
M  & rhyth{\bf m} \\
$\sim$  &  butt{\bf on} \\ \hline
\end{tabular}
\begin{tabular}{|c|l|} 
\multicolumn{2}{c}{\bf Vowels}\\ \hline
{\bf ASCII} & {\bf Example}  \\ \hline \hline
I  & b{\bf i}t \\
E  & b{\bf  e}t \\
\& & b{\bf  a}t \\
A  & b{\bf  u}t \\
a  & h{\bf  o}t \\
O  & l{\bf  aw} \\
U  & p{\bf  u}t \\
6  & h{\bf  e}r \\
i  & b{\bf ee}t \\
e  & b{\bf ai}t \\
u  & b{\bf oo}t\\
o  & b{\bf oa}t \\
9  & b{\bf uy} \\
Q  & b{\bf ou}t \\
7  & b{\bf oy} \\\hline
\end{tabular}
\begin{tabular}{|c|l|}
\multicolumn{2}{c}{\bf Vowel + r}\\ \hline
{\bf ASCII} & {\bf Example}  \\ \hline \hline
3  & b{\bf ir}d\\
R  & butt{\bf er} \\
\# & {\bf ar}m \\
\% & h{\bf or}n \\
*  & {\bf air} \\
(  & {\bf ear} \\
)  & l{\bf ure} \\ \hline
\end{tabular}
\end{singlespace}
\vspace{0.5cm}

\starttwocolumn
\section*{Acknowledgments}

The author thanks Michael Brent for stimulating his interest in the area.
Thanks are also due to Koryn Grant for verifying the results presented
here.  Eleanor Olds Batchelder and Andreas Stolcke gave many
constructive comments and useful pointers in preparing a revised version of
this paper.  The ``{\em Damn British}''anecdote is due to Robert
Linggard.  Judith Lee at SRI edited the manuscript to remove many
typographical and stylistic errors.  In addition, the author is very
grateful to anonymous reviewers of an initial version, especially ``B'' and
``D'', who helped significantly by way of encouragement and constructive
criticism.

\bibliography{venkataraman-00}

\begin{thebibliography}{}

\bibitem[\protect\citename{Ando and Lee}1999]{Ando:USS99}
Ando, R.~K. and Lillian Lee.
\newblock 1999.
\newblock Unsupervised statistical segmentation of {Japanese Kanji} strings.
\newblock Technical Report TR99-1756, Cornell University, Ithaca, NY.

\bibitem[\protect\citename{Batchelder}1997]{Batchelder:CEU97}
Batchelder, Eleanor~Olds.
\newblock 1997.
\newblock {\em Computational evidence for the use of frequency information in
  discovery of the infant's first lexicon}.
\newblock Unpublished {PhD} dissertation, {City University of New York}, New
  York, NY.

\bibitem[\protect\citename{Bernstein-Ratner}1987]{Bernstein:PPC87}
Bernstein-Ratner, N.
\newblock 1987.
\newblock The phonology of parent child speech.
\newblock In K.~Nelson and A.~van Kleeck, editors, {\em Children's Language},
  volume~6. Erlbaum, Hillsdale, NJ.

\bibitem[\protect\citename{Brent}1999]{Brent:EPS99}
Brent, Michael~R.
\newblock 1999.
\newblock An efficient probabilistically sound algorithm for segmentation and
  word discovery.
\newblock {\em {M}achine {L}earning}, 34:71--105.

\bibitem[\protect\citename{Brent and Cartwright}1996]{Brent:DRP96}
Brent, Michael~R. and Timothy~A. Cartwright.
\newblock 1996.
\newblock Distributional regularity and phonotactics are useful for
  segmentation.
\newblock {\em Cognition}, 61:93--125.

\bibitem[\protect\citename{Christiansen, Allen, and
  Seidenberg}1998]{Christiansen:LSS98}
Christiansen, M.~H., J.~Allen, and M.~Seidenberg.
\newblock 1998.
\newblock Learning to segment speech using multiple cues: {A} connectionist
  model.
\newblock {\em Language and cognitive processes}, 13:221--268.

\bibitem[\protect\citename{Cutler and Carter}1987]{Cutler:PSI87}
Cutler, Anne and D.~M. Carter.
\newblock 1987.
\newblock Predominance of strong initial syllables in the {E}nglish vocabulary.
\newblock {\em Computer speech and language}, 2:133--142.

\bibitem[\protect\citename{de Marcken}1995]{deMarcken:UAL95}
de~Marcken, C.
\newblock 1995.
\newblock Unsupervised acquisition of a lexicon from continuous speech.
\newblock Technical Report AI Memo No. 1558, Massachusetts Institute of
  Technology, Cambridge, MA.

\bibitem[\protect\citename{Elman}1990]{Elman:FST90}
Elman, J.~L.
\newblock 1990.
\newblock Finding structure in time.
\newblock {\em Cognitive Science}, 14:179--211.

\bibitem[\protect\citename{Haussler}1988]{Haussler:QIB88}
Haussler, David.
\newblock 1988.
\newblock Quantifying inductive bias: {AI} learning algorithms and {V}aliant's
  learning framework.
\newblock {\em {Artificial Intelligence}}, 36:177--221.

\bibitem[\protect\citename{Jelinek}1997]{Jelinek:SMS97}
Jelinek, F.
\newblock 1997.
\newblock {\em Statistical methods for speech recognition}.
\newblock MIT Press, Cambridge, MA.

\bibitem[\protect\citename{Jusczyk}1997]{Jusczyk:DSL97}
Jusczyk, Peter~W.
\newblock 1997.
\newblock {\em The discovery of spoken language}.
\newblock MIT Press, Cambridge, MA.

\bibitem[\protect\citename{Jusczyk, Cutler, and Redanz}1993]{Jusczyk:PDS93}
Jusczyk, Peter~W., Anne Cutler, and N.~Redanz.
\newblock 1993.
\newblock Preference for predominant stress patterns of {E}nglish words.
\newblock {\em Child Development}, 64:675--687.

\bibitem[\protect\citename{Jusczyk and Hohne}1997]{Jusczyk:IMS97}
Jusczyk, Peter~W. and E.~A Hohne.
\newblock 1997.
\newblock Infants' memory for spoken words.
\newblock {\em Science}, 27:1984--1986.

\bibitem[\protect\citename{Katz}1987]{Katz:EPF87}
Katz, Slava~M.
\newblock 1987.
\newblock Estimation of probabilities from sparse data for the language model
  component of a speech recognizer.
\newblock {\em {IEEE Transactions on Acoustics, Speech and Signal Processing}},
  ASSP-35(3):400--401.

\bibitem[\protect\citename{MacWhinney and Snow}1985]{MACWHINNEY:CLD85}
MacWhinney, Brian and C.~Snow.
\newblock 1985.
\newblock The child language data exchange system.
\newblock {\em Journal of {C}hild {L}anguage}, 12:271--296.

\bibitem[\protect\citename{Mattys and Jusczyk}1999]{Mattys:PPE99}
Mattys, Sven~L. and Peter~W. Jusczyk.
\newblock 1999.
\newblock Phonotactic and prosodic effects on word segmentation in infants.
\newblock {\em Cognitive psychology}, 38:465--494.

\bibitem[\protect\citename{Olivier}1968]{Olivier:SGL68}
Olivier, D.~C.
\newblock 1968.
\newblock {\em Stochastic grammars and language acquisition mechanisms}.
\newblock Unpublished {PhD} dissertation, Harvard University, Cambridge, MA.

\bibitem[\protect\citename{Saffran, Newport, and Aslin}1996]{Saffran:WSR96}
Saffran, Jennifer~R., E.~L. Newport, and R.~N. Aslin.
\newblock 1996.
\newblock Word segmentation: {T}he role of distributional cues.
\newblock {\em Journal of {M}emory and {L}anguage}, 35:606--621.

\bibitem[\protect\citename{Viterbi}1967]{Viterbi:EBC67}
Viterbi, Andrew~J.
\newblock 1967.
\newblock Error bounds for convolutional codes and an asymptotically optimal
  decoding algorithm.
\newblock {\em {IEEE} {T}ransactions on {I}nformation {T}heory},
  IT-13:260--269.

\bibitem[\protect\citename{Witten and Bell}1991]{Witten:ZFP91}
Witten, Ian~H. and Timothy~C. Bell.
\newblock 1991.
\newblock The zero-frequency problem: Estimating the probabilities of novel
  events in adaptive text compression.
\newblock {\em {IEEE} {T}ransactions on {I}nformation {T}heory},
  37(4):1085--1091.

\bibitem[\protect\citename{Zimin and Tseng}1993]{Zimin98:CTS93}
Zimin, W. and G.~Tseng.
\newblock 1993.
\newblock Chinese text segmentation for text retrieval problems and
  achievements.
\newblock {\em JASIS}, 44(9):532--542.

\end{thebibliography}

\end{document}